# Beyond Context to Cognitive Appraisal: Emotion Reasoning as a Theory of Mind Benchmark for Large Language Models

**Gerard Christopher Yeo[1], Kokil Jaidka[1]**

[1]NUS Centre for Trusted Internet and Community, National University of Singapore

e0545159@u.nus.edu, jaidka@nus.edu.sg

## Abstract

Datasets used for emotion recognition tasks typically contain overt cues that can be used in predicting the emotions expressed in a text. However, one challenge is that texts sometimes contain covert contextual cues that are rich in affective semantics, which warrant higher-order reasoning abilities to infer emotional states, not simply the emotions conveyed. This study advances beyond surface-level perceptual features to investigate how large language models (LLMs) reason about others' emotional states using contextual information, within a Theory-of-Mind (ToM) framework. Grounded in Cognitive Appraisal Theory, we curate a specialized ToM evaluation dataset[1] to assess both forward reasoning—from context to emotion—and backward reasoning—from emotion to inferred context. We showed that LLMs can reason to a certain extent, although they are poor at associating situational outcomes and appraisals with specific emotions. Our work highlights the need for psychological theories in the training and evaluation of LLMs in the context of emotion reasoning.

## 1 Introduction

Humans are naturally adept at reasoning about other people's mental states such as beliefs, goals, and intentions (Tomasello et al., 2005; Zaki and Ochsner, 2011). These social-cognitive reasoning abilities enable individuals to navigate complex social situations and facilitate interpersonal communication (Imuta et al., 2016). Recent research has explored whether LLMs possess such abilities to infer cognitive states of other agents (e.g. theory-of-mind (ToM)) (Kim et al., 2023; Jung et al., 2024), however, LLMs' ability to reason about one of the most essential components of human experiences—*emotional states*—is not well studied.

[1]Data is publicly available on https://github.com/GerardYeo/ToMEmoReason.git

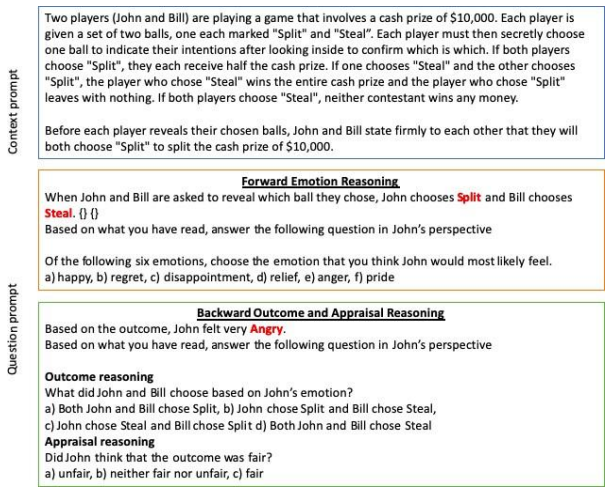

Figure 1: Overview of emotion reasoning in LLMs in the current study. The vignette is an example of the Prisoner's Dilemma problem contextualised in a high-stake gameshow (Golden Balls). The curly braces in the context prompt consist of the appraisal prompts.

Studies in affective computing commonly engage in emotion recognition tasks where the goal is to predict the most representative emotion in a text (Deng and Ren, 2021). Although recent transformer models can represent contextual cues to a certain extent, it is not clear whether these models are reasoning about agents' emotional states or they are merely using emotional words in the prediction. *Emotion reasoning* is conceptually different from *emotion recognition* where the former incorporates abstract contextual information to reason about emotions while the latter is typically based on perceptual features (i.e. emotional words).

Reasoning about other people's emotions involves a set of cognitive processes and numerous psychological studies have adapted an intuitive ToM framework to understand these processes (Houlihan et al., 2022; Ong et al., 2015). For instance, given that John received a poor grade on an exam, an observer could reason that John is feeling sad. In addition to inferring emotions based on a situation (forward reasoning), people also perform backward reasoning (seeing John crying) where they infer possible situations that elicit particular emotions (De Melo et al., 2014; Hareli and Hess, 2010). These and other paradigms are discussed as part of cognitive psychology, where empirical findings that such emotion reasoning in humans can be understood through the framework of dual-process theories, which posit two distinct modes of thinking: System 1, which is fast, automatic, and affect-driven, and System 2, which is slower, deliberate, and effortful (Kahneman, 2011; Evans, 2008). Emotion reasoning likely requires a shift from System 1's pattern-based recognition to System 2's context-sensitive inference, with accurate cognitive inference serving as a crucial intermediate step. For LLMs, this process might involve using contextual cues to simulate the evaluative processes that guide human emotion inference, rather than simply relying on surface-level emotional features. Yet, which paths do LLMs choose for emotional reasoning?

To answer this question, this study examines whether LLMs can extend beyond basic emotion recognition to infer third-person emotions based on contextual cues, and infer outcomes based on emotions. We developed a dataset comprising vignettes that are systematically manipulated according to cognitive appraisal theory. Cognitive Appraisal Theory, which is one of the primary theoretical frameworks for understanding emotional experiences, posits that emotions are elicited from subjective interpretation of events along a set of dimensions (Ellsworth and Scherer, 2003; Moors et al., 2013; Yeo and Ong, 2023). Although certain situations are known to reliably elicit specific emotions in people (e.g. death of a loved one results in sadness), Cognitive Appraisal Theory emphasizes that it is the evaluation of the situation, rather than the situation per se, that results in emotional reactions. For instance, if an individual appraises a situation as obstructing their personal goals (goal inconduciveness), attributes responsibility to another person (other-accountability), and perceives the situation as unjust (unfairness), they are likely to experience anger. Conversely, if another individual appraises the same situation differently, a different emotional response is likely to be elicited. Observers inferring another agent's emotions engage in a comparable cognitive process, drawing on their understanding of the agent's goals and desires, as well as how the agent is likely to evaluate the situation (Ong et al., 2015; Skerry and Saxe, 2015). That is, by considering the alignment between the situation and the agent's goals—alongside intuitive emotion knowledge (e.g., unfairness typically elicits anger)—observers make inferences about the agent's emotional state. Our research builds on prior work investigating ToM in LLMs, extending it to third-person emotion reasoning through the lens of Cognitive Appraisal Theory (see Figure 1). We designed vignettes that vary in their linguistic descriptions of outcomes and cognitive appraisals while minimizing explicit emotional features (Roseman, 1991). Our theoretical contributions are as follows:

- We have created a new ToM task to elucidate how state-of-the-art LLMs reason about emotions in other agents based on manipulated vignettes, grounded in Cognitive Appraisal Theory (Ellsworth and Scherer, 2003).
- We have created subtasks that provide evidence for the System-1 and System-2 reasoning mechanisms applied by LLMs.
- We have curated a novel emotion reasoning evaluation dataset and evaluate LLMs' emotion reasoning abilities.

## 2 Evaluation Dataset, Prompts, and Tasks

We constructed our evaluation dataset based on the Prisoner's Dilemma game, adapting stimuli from previous psychological studies into a text-based format (see Figure 1; Houlihan et al. (2022, 2023)). Despite its simplicity, this context provides a highly relevant framework for examining how LLMs infer others' emotions. Having vignettes with clearly defined outcomes based on the decisions of the characters, the dataset enables a systematic investigation of emotion reasoning processes.

Our evaluation dataset comprises two primary tasks: (a) forward emotion reasoning and (b) backward outcome and appraisal reasoning. In the forward emotion reasoning task, LLMs are prompted to infer the emotions of a target agent in a given vignette, based on descriptions of the outcomes and the agent's cognitive appraisals. This task's evaluation dataset includes three distinct scenarios representing the Prisoner's Dilemma in different contexts: (i) the Split or Steal game show, (ii) a business deal, and (iii) relationship commitment (see Appendix A). These contexts were selected for their relevance to high-stakes decision-making domains. The evaluation prompts were systematically designed to manipulate the following key variables: (1) outcome choices, (2) cognitive appraisal of fairness, and (3) accountability.

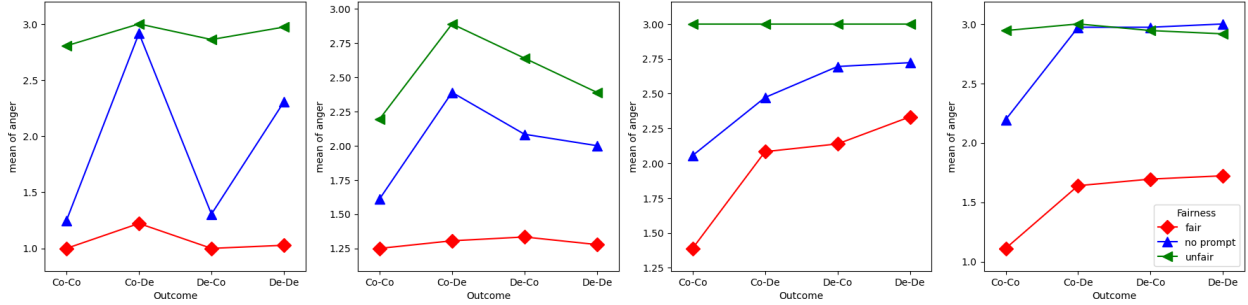

Figure 2: Interaction plots between outcomes and fairness in influencing mean anger ratings for all models.

To account for the sensitivity of LLM responses, we rephrased each vignette in three different ways using GPT-4. After being presented with the designed vignettes, LLMs were required to answer an appraisal and emotion questionnaire (see Appendix A) These questions were adapted from previous studies on cognitive appraisals (Frijda, 1987; Scherer, 1997). In total, the forward reasoning task comprises 432 vignettes.

For the backward reasoning task, we assessed the ability of LLMs to infer the outcomes and cognitive appraisals when provided with a vignette that explicitly states only the target agent's emotion. This task utilized the same prompts as the forward reasoning task, but with the outcome and appraisals omitted (see Appendix B). Each prompt was systematically manipulated by varying the six possible emotions experienced by the target agent. After being presented with the scenario, LLMs were required to answer a series of questions regarding the outcome chosen by the agent and the agent's appraisal of the situation. In total, the backward reasoning task comprises 150 evaluation prompts.

We used the curated vignettes and questions as input prompts (see Appendix F for dataset statistics) for four state-of-the-art LLMs, namely Mistral 7B (Mistral-7B-Instruct-v0.3) (Jiang et al., 2023), Llama 3.1 (Llama-3.1-8b-Instruct) (Dubey et al., 2024), Gemma 7b (gemma-7b-Instruct) (Team et al., 2024), and OpenAI o3 (o3-mini). The models then answer the questions in a zero-shot manner. Default parameters are used for all the models during inference.

## 3 Results

### 3.1 Forward Emotion Reasoning

For each of the designed vignettes with differing outcomes and appraisals, we use the theoretical appraisal-emotion relationship that cognitive appraisal theory has proposed as the "ground truth" emotion label (Frijda et al. (1989); Roseman and Smith (2001); see Appendix D). These outcome–appraisal–emotion mappings are not intended to represent fixed or universal ground truths, but are instead grounded in well-established empirical findings from cognitive appraisal theory, specifically within the context of the Prisoner's Dilemma (Houlihan et al., 2023). They serve as a theoretically informed baseline for assessing the reasoning patterns of LLMs in relation to human psychological expectations. When prompted with a specific vignette, the LLMs are tasked with inferring the most representative emotion—selected from a set of six emotions (anger, disappointment, joy, pride, regret, relief)—that the target agent would likely experience. These inferred emotions are then compared against the theory-informed emotion labels to assess alignment.

For this six-way emotion classification task, Gemma 7b outperformed the rest by obtaining an accuracy score of 57.9%, followed by OpenAI's o3 (55.6%), Mistral 7B (54.0%), and Llama 3.1 (52.0%). This suggests that LLMs perform moderately well (about 35 - 40% above chance), moderately consistent to what cognitive appraisal theories have proposed.

One possible explanation for Gemma-7B outperforming OpenAI's o3-mini is that o3-mini has been explicitly optimized for logical reasoning tasks, particularly in structured domains such as science and mathematics. As a result, it may deprioritize the types of appraisal patterns related to emotional reasoning that are central to the current task. In con-trast, Gemma-7B has been trained on a broader and more diverse corpus, encompassing a wide range of domains. This likely makes it better suited for general natural language understanding and reasoning, rather than for narrowly defined problem-solving tasks (Salido et al., 2025). Consequently, Gemma-7B may be more inclined to rely on surface-level heuristics or pattern recognition learned during pretraining—patterns that potentially align more closely with the emotional structures observed in the Prisoner's Dilemma scenario.

Beyond identifying a single representative emotion, the LLMs also responded to multiple questions regarding the intensity of specific emotions (see Figure 7 in Appendix A). This approach accounts for the possibility that the target agent may experience more than one emotion simultaneously, allowing for a more nuanced analysis of emotional inference. To examine how various outcomes and appraisals influence emotional intensity responses, we conducted two-and three-way Analysis of Variance (ANOVA) for each model and emotion rating. While the three-way interaction effects were not statistically significant across models and emotions ($F$s(18, 384) < 1.32, $p$s > .05), suggesting that emotional intensities are not determined by how LLMs integrate outcome and appraisal information, most two-way interactions between outcome and the appraisal of fairness were statistically significant ($F$s(6, 384) > 6.37, $p$s < .05). This finding indicates that LLMs primarily rely on outcome information and fairness appraisals when reasoning about an agent's emotions, while the appraisal of accountability appears to play a minimal role.

Figure 2 illustrates the interaction between outcome and the appraisal of fairness in influencing the mean anger scores (see Appendix C for the results for accountability). The plot shows that when the target agent chooses to "cooperate" while the other agent chooses to "defect," anger scores tend to be higher. However, in cases where the vignette explicitly describes an appraisal of unfairness, certain LLMs (e.g., Gemma) consistently generate high anger intensity scores regardless of the outcome. This suggests that while LLMs reason about emotions in a manner consistent with appraisal theory—where fairness appraisals are closely linked to anger (Scherer, 1997)—they tend to prioritize specific inferred appraisals over others. For instance, unfairness appears to be a more dominant factor in predicting anger than goal conduciveness.

Moreover, this pattern is indicative of System 1 thinking, wherein models rely on heuristic-based associations—such as consistently linking anger with perceptions of unfairness—regardless of contextual nuances. This suggests the use of fast, affect-driven processing mechanisms rather than deliberate, abstract reasoning.

## 3.2 Backward Emotion Reasoning

### 3.2.1 Reasoning about Contextual Outcomes

Now, we analyze how LLMs associate specific emotions to specific outcomes. The heatmaps in Figure 3 display the results of the backward outcome reasoning task for each model. These visualizations show the frequency with which the LLMs assign specific emotion-outcome pairs. A high frequency for a given emotion-outcome pair indicates that the LLMs consistently associate that particular emotion with a specific type of outcome.

In general, anger, disappointment, and regret are associated with the outcome in which the target agent chooses "Cooperate" ('Co') while the other agent chooses "Defect" ('De'). This pattern is most pronounced in the o3 and Gemma models. According to appraisal theories, the "Co-De" outcome is closely linked to anger due to the perceived unfairness when both agents initially agreed to cooperate, yet one defects. However, the results reveal that LLMs associate both anger and disappointment with the "Co-De" outcome. This suggests that, beyond relying on fairness appraisals, LLMs may also use other mechanisms—such as assessing the pleasantness of the situation—to infer emotions. Interestingly, the "De-De" condition was rarely endorsed by the LLMs, despite the outcome being equally unfavorable as the "Co-De" condition. We discuss these findings in the following section.

### 3.2.2 Reasoning about Cognitive Appraisals

Finally, we evaluated LLMs' ability to reason about cognitive appraisals that could evidence System 2 emotion reasoning. For each model, we conducted a series of univariate ANOVAs, with the manipulated emotions serving as the grouping variable and a set of appraisal questions as the dependent variables to examine whether there are systematic differences in appraisal ratings between emotions that are consistent with appraisal theories. We found statistically significant differences in appraisal ratings between emotions across all models, except for Llama 3 in the accountability-other ($F$(5, 144) = 1.14, $p$ > .05) and accountability-self ($F$(5, 144) = 1.78, $p$ > .05) appraisals. This suggests that Llama

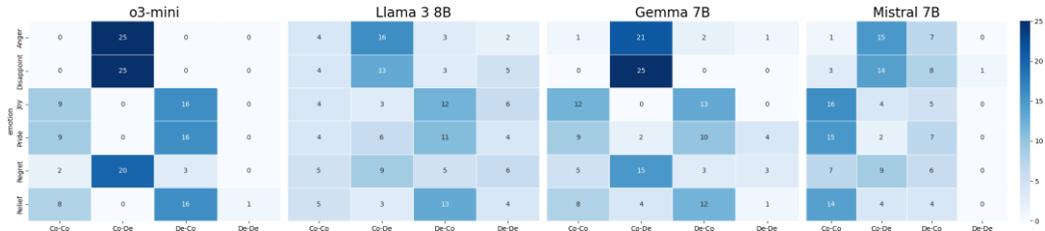

Figure 3: Heatmap of the frequencies for the outcomes (x-axis) when models are prompted with target agent having different emotions (y-axis). 'Co' and 'De' refers to Cooperate and Defect, respectively.

3 does not recognize the systematic differences between emotions in relation to accountability-related appraisals. Additionally, all models were able to associate specific emotions with appraisals of goal conduciveness. For instance, the goal conduciveness scores for positive emotions, such as joy, were higher than for negative emotions, such as anger, in the o3 model ($M_{diff} = 2.0$, $p < .05$).

Upon closer examination of accountability-related appraisals, although significant differences between emotions were observed for the o3, Gemma, and Mistral models, the models still struggle to reliably associate specific emotions with corresponding appraisals.

## 4 Discussion and Conclusion

The current work builds on prior research examining ToM in LLMs with a focus on emotional inference, and extends psychological studies that elucidate the human emotion inference process (Houlihan et al., 2023; Ong et al., 2015) to the domain of text-based emotion inference in LLMs. Overall, our findings suggest that while LLMs can reason to some degree, they struggle to accurately associate situational outcomes and appraisals with specific emotions. The results of both the forward and backward reasoning tasks suggest that LLMs predominantly adopt System 1-like strategies. In the forward reasoning task, for instance, anger ratings were consistently elevated in scenarios with perceived unfairness, regardless of outcome—a heuristic shortcut that bypasses the nuanced integration of contextual factors indicative of System 2 engagement.

The backward reasoning results further underscore this System 1 dominance, as LLMs frequently conflated distinct but similar emotions, such as anger and disappointment, which require subtle distinctions based on appraisal dimensions like accountability and goal conduciveness. This difficulty suggests that the models possess limited capacity to emulate System 2-like processes, which entail abstract, theory-informed reasoning grounded in event appraisals and a nuanced understanding of social context (Frijda et al., 1989; Ong et al., 2015). Such processes require the integration of multiple contextual cues to generate a coherent and holistic interpretation of a target's emotional state, rather than relying solely on a single modality or superficial features.

Taken together, these findings indicate that LLMs' emotional reasoning predominantly reflects heuristic-driven, pattern-matching processes consistent with System 1, with limited evidence of System 2-like context integration. Future research may benefit from designing training protocols that encourage more abstract, rule-based inference mechanisms to better approximate the cognitive pathways underlying human emotional reasoning.

## Limitations

This study investigates emotion reasoning within a constrained context—the Prisoner's Dilemma game. However, real-world conversations and social interactions are often more complex. The primary objective of this research is to assess whether LLMs can engage in both forward and backward emotion reasoning using systematically designed vignettes. Future research could build upon these findings by exploring emotion reasoning in more naturalistic conversations and real-world textual contexts.

A comprehensive investigation of emotion reasoning may require considering a broader range of appraisal dimensions, as individuals could potentially utilize numerous relevant appraisals to evaluate third-person reasoning. In the context of studying the relationship between appraisal and emotions, our choice of appraisal dimensions was therefore determined by prior research which has demonstrated that the set of emotions examined in our study are significantly associated with 3 key appraisal dimensions (Yeo and Ong, 2023). Future research could extend our work by incorporating a broader range of appraisal dimensions to further explore the complexities of emotion reasoning.

Furthermore, even within the specific context of our study—namely, the Prisoner's Dilemma paradigm, which involves distinct outcome choices—prior research has reported that human participants typically use the appraisals of goal conduciveness, fairness, and accountability to reason about the quality and intensity of emotions experienced by a target agent (Houlihan et al., 2023). For instance, goal conduciveness pertains to whether the target agent receives the mutual reward for cooperation or suffers a loss due to defection. Fairness relates to whether both agents honored their agreement to cooperate or whether one exploited the other. Accountability concerns whether the emotional response is directed toward the other agent's betrayal or attributed to the target agent's own choice to trust. In this framework, goal conduciveness can help distinguish between broadly positive and negative emotional outcomes, while fairness and accountability appraisals can help differentiate emotions such as gratitude versus relief, or anger versus guilt.

## Ethics Statement

Reasoning about emotions is a complex phenomenon which is highly dependent on the context, situation, and culture. Moreover, the issue of multiple emotions (i.e. a person can experience multiple different emotions at the same time) further complicates the issue of emotion inference. Users of models that engage in emotion recognition and reasoning tasks in real-life cases should always account for the uncertainty of predictions.

All data was curated manually with theoretical grounding in cognitive appraisal theory and so there is no ethical concern surrounding the data collection process.

**Acknowledgments.** This work is supported by the Ministry of Education, Singapore, under its MOE AcRF TIER 3 Grant (MOE-MOET32022-0001).

## A  Forward Reasoning Vignettes and Questionnaire

Figure 4, 5, and 6 present the designed vignettes for the forward reasoning task. LLMs are prompted with these vignettes before answering a set of emotion and appraisal questions. Figure 7 presents the questions LLMs answer after being prompted with the vignettes.

Figure 4: Vignette for the Split or Steal Game Show for the forward reasoning task. The italicized word in curly brackets is the outcome and appraisal phrase that is manipulated.

Figure 5: Vignette for the Business Deal scenario for the forward reasoning task. The italicized word in curly brackets is the outcome and appraisal phrase that is manipulated.

Figure 6: Vignette for the Relationship Commitment scenario for the forward reasoning task. The italicized word in curly brackets is the outcome and appraisal phrase that is manipulated.

## B  Backward Reasoning Vignettes and Questionnaire

Figure 8, 9, and 10 present the designed vignettes for the backward reasoning task. LLMs are

1. Of the following six emotions, choose the emotion that you think {agent_t} would most likely feel.
    a) happy, b) regret, c) disappointment, d) relieved, e) anger, f) pride
2. How happy did {agent_t} feel?
    a) Not happy, b) Slightly happy, c) Very happy
3. How regretful did {agent_t} feel?
    a) Not regretful, b) Slightly regretful, c) Very regretful
4. How disappointed did {agent_t} feel?
    a) Not disappointed, b) Slightly disappointed, c) Very disappointed
5. How relieved did {agent_t} feel?
    a) Not relieved, b) Slightly relieved, c) Very relieved
6. How angry did {agent_t} feel?
    a) Not angry, b) Slightly angry, c) Very angry
7. How proud did {agent_t} feel?
    a) Not proud, b) Slightly proud, c) Very proud
8. Did {agent_t} think that he was mainly responsible for the outcome?
    a) {agent_t} thinks that he was not responsible
    b) {agent_t} thinks that he was slightly responsible
    c) {agent_t} thinks that he was very responsible
9. Did {agent_t} think that {agent_o} was mainly responsible for the outcome?
    a) {agent_t} thinks that {agent_o} was not responsible
    b) {agent_t} thinks that {agent_o} was slightly responsible
    c) {agent_t} thinks that {agent_o} was very responsible
10. Did {agent_t} think that the outcome was fair?
    a) {agent_t} thinks the outcome was unfair
    b) {agent_t} thinks the outcome was neither fair nor unfair
    c) {agent_t} thinks the outcome was fair
11. Did the outcome meet {agent_t}'s goals?
    a) {agent_t} thinks that his goals were not met
    b) {agent_t} thinks that his goals were neither met nor not met
    c) {agent_t} thinks that his goals were met

Figure 7: Questionnaire for the forward reasoning task. $agent_t$ and $agent_o$ represent the target and other agent, respectively.

prompted with these vignettes before answering a set of outcome and appraisal questions. Figure 11 presents the questions LLMs answer after being prompted with the vignettes.

**Split or Steal Game Show**
Two players (John and Bill) are playing a game that involves a cash prize of $10,000. Each player is given a set of two balls, one each marked "Split" and "Steal". Each player must then secretly choose one ball to indicate their intentions after looking inside to confirm which is which. If both players choose "Split", they each receive half the cash prize. If one chooses "Steal" and the other chooses "Split", the player who chose "Steal" wins the entire cash prize and the player who chose "Split" leaves with nothing. If both players choose "Steal", neither contestant wins any money.

Before each player reveals their chosen balls, John and Bill state firmly to each other that they will both choose "Split" to split the cash prize of $10,000.

When John and Bill are asked to reveal which ball they chose. Based on the outcome, John felt very {emotion}.
Based on what you have read, answer the following question in John's perspective.

emotion = [angry, happy, disappointed, proud, relieved, regretful]

Figure 8: Vignette for the Split or Steal Game Show for the backward reasoning task. The italicized word in curly brackets is the emotion that is manipulated.

## C Forward Reasoning Accountability results

With regard to the appraisal of accountability, LLMs do not appear to utilize this appraisal information when reasoning about emotions. Specifically, in the case of anger (see Figure 12), we expect that anger intensity ratings for the "Cooperate-Defect" condition would be highest when the vignette indicated that the other agent was primarily

**Business Deal**
Alex and Peter, the CEOs of two rival firms, have an opportunity to collaborate on a groundbreaking AI venture that could generate $10 million in revenue. Each must make an independent decision: Cooperate or Compete. If both choose to cooperate, they establish a partnership and share the profits equally, earning $5 million each. If one cooperates while the other competes, the competing CEO dominates the market, claiming the entire $10 million, while the cooperative CEO gets nothing. If both compete, they enter a fierce price war, leading to zero profit for both.

Before making their choices official, Alex and Peter agree to cooperate and split the profits.

However, when their final decisions are revealed, Alex felt very {emotion}.
Based on Alex's emotion, select the most accurate multiple-choice answer that reflects Alex's perspective.

emotion = [angry, happy, disappointed, proud, relieved, regretful]

Figure 9: Vignette for the Business Deal scenario for the forward reasoning task. The italicized word in curly brackets is the emotion that is manipulated.

**Relationship Commitment**
Kelvin and Brenda have been dating for a while and now face a crucial decision: Commit or Ghost. If both commit, they build a strong relationship and eventually get married. If one commits while the other ghosts, the committed person will be wasting his or her time, missing out on a meaningful connection, while the ghoster avoids responsibility and keeps their options open. If both ghost, they waste time and miss out on a meaningful connection.

Before making their choices official, Kelvin and Brenda agree to commit and hopefully get married.

However, when their final decisions are revealed, Kelvin felt very {emotion}.
Based on Kelvin's emotion, select the most accurate multiple-choice answer that reflects Kelvin's perspective

emotion = [angry, happy, disappointed, proud, relieved, regretful]

Figure 10: Vignette for the Relationship Commitment scenario for the forward reasoning task. The italicized word in curly brackets is the emotion that is manipulated.

1. What did {agent_t} and {agent_o} choose based on {agent_t}'s emotion?
    a) Both {agent_t} and {agent_o} chose {cooperate}
    b) {agent_t} chose {cooperate} and {agent_o} chose {defect}
    c) {agent_t} chose {defect} and {agent_o} chose {cooperate}
    d) Both {agent_t} and {agent_o} chose {defect}
2. Did {agent_t} think that he was mainly responsible for the outcome?
    a) {agent_t} thinks that he was not responsible
    b) {agent_t} thinks that he was slightly responsible
    c) {agent_t} thinks that he was very responsible
3. Did {agent_t} think that {agent_o} was mainly responsible for the outcome?
    a) {agent_t} thinks that {agent_o} was not responsible
    b) {agent_t} thinks that {agent_o} was slightly responsible
    c) {agent_t} thinks that {agent_o} was very responsible
4. Did {agent_t} think that the outcome was fair?
    a) {agent_t} thinks the outcome was unfair
    b) {agent_t} thinks the outcome was neither fair nor unfair
    c) {agent_t} thinks the outcome was fair
5. Did the outcome meet {agent_t}'s goals?
    a) {agent_t} thinks that his goals were not met
    b) {agent_t} thinks that his goals were neither met nor not met
    c) {agent_t} thinks that his goals were met

Figure 11: Questionnaire for the backward reasoning task. $agent_t$ and $agent_o$ represent the target and other agent, respectively. *cooperate* and *defect* could refer to different terms in different vignettes. For example these refers to 'split' and 'steal' in the game show scenario, respectively.

responsible for the outcome. However, this was not supported by our findings.

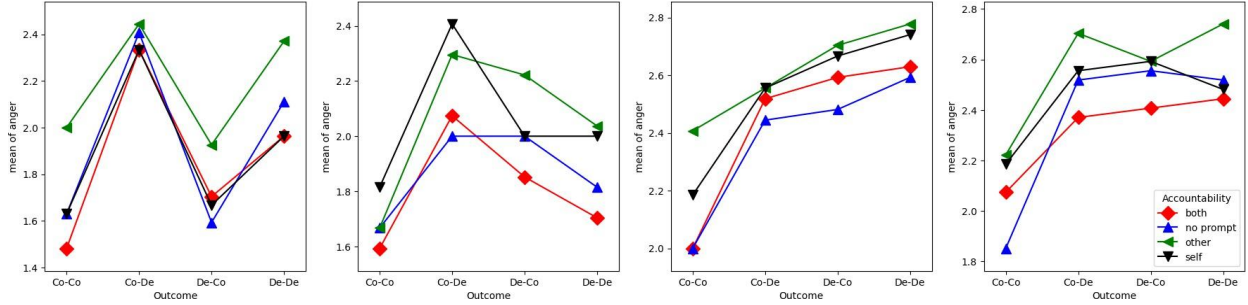

Figure 12: Interaction plots between outcomes and accountability in influencing mean anger ratings for all models.

| Emotion | Target agent | Other agent |
|---|---|---|
| Anger | Co | De |
| Joy | Co/De | Co |
| Disappointment | Co/De | De |
| Pride | De | Co |
| Regret | Co | Co/De |
| Relief | Co | Co |

Table 1: Theoretical predictions of outcomes with respect to specific emotions. "Co" and "De" refer to cooperate and defect, respectively.

| Task/Domain | Mean | Min | Max |
|---|---|---|---|
| *Forward* | | | |
| Split-Steal | 170.4 | 141 | 209 |
| Business deal | 173.0 | 159 | 182 |
| Relationship | 156.4 | 143 | 168 |
| *Backward* | | | |
| Split-Steal | 151.2 | 128 | 206 |
| Business deal | 110.4 | 79 | 157 |
| Relationship | 102.0 | 81 | 141 |
| Climate | 89.2 | 70 | 127 |
| Cybersecurity | 151.2 | 70 | 100 |

Table 2: The number of words in the dataset, categorized by task and domain.

## D Theoretical predictions outcome-emotion classification

Table 1 presents the theoretical hyptheses between outcomes and emotions (Houlihan et al., 2022).

## E Manipulation Checks for Forward Reasoning Task

We first conducted manipulation checks to determine whether the LLMs understood the prompts and the questions. We segmented each appraisal score into two groups based on the experimental manipulation (i.e. high vs. low) and then compared the appraisal scores between these two groups using independent sample t-test test. For example, for the prompt of fairness, since we manipulated the prompt to elicit both high and low fairness scores, the fairness scores for the high group are compared with the low group. If the manipulation of prompts is effective, it should result in the appropriate appraisal scores (i.e. low fairness score for the low group and vice versa).

From our results, the scores of all models are consistent with the manipulated vignettes, indicating that the LLMs understood the vignettes and questions.

## F Dataset Statistics

Table 2 summarizes the statistics of the constructed evaluation dataset across different tasks and domains. The dataset comprises 432 instances for the forward reasoning task and 150 instances for the backward reasoning task. For both tasks, the instances are evenly distributed across the domains.